# Intersection focused Situation Coverage-based Verification and Validation Framework for Autonomous Vehicles Implemented in CARLA

Zaid Tahir[1(✉)] and Rob Alexander[2]

[1,2] Assuring Autonomy International Programme, Department of Computer Science, University of York, York, United Kingdom.
[1]Boston University, ECE Department, United States.
[1][zaid.butt.tahir@gmail.com, zaidt@bu.edu], [2]rob.alexander@york.ac.uk

**Abstract.** Autonomous Vehicles (AVs) i.e., self-driving cars, operate in a safety-critical domain, since errors in the autonomous driving software can lead to huge losses. Statistically, road intersections which are a part of the AVs operational design domain (ODD), have some of the highest accident rates. Hence, testing AVs to the limits on road intersections and assuring their safety on road intersections is pertinent, and thus the focus of this paper. We present a situation coverage-based (SitCov) AV-testing framework for the verification and validation (V&V) and safety assurance of AVs, developed in an open-source AV simulator named CARLA. The SitCov AV-testing framework focuses on vehicle-to-vehicle (V2V) interaction on a road intersection under different environmental conditions and intersection configuration situations (start/goal locations), using situation coverage criteria for automatic test suite generation for safety assurance of AVs. We have developed an ontology for intersection situations, and used it to generate a situation hyperspace i.e., the space of all possible situations arising from that ontology. For the evaluation of our SitCov AV-testing framework, we have seeded multiple faults in our ego AV, and compared situation coverage-based and random situation generation. We have found that both generation methodologies trigger around the same number of seeded faults, but the situation coverage-based generation tells us a lot more about the weaknesses of the autonomous driving algorithm of our ego AV, especially in edge-cases. Our code is publicly available online and since the simulation software (CARLA) is open-source, anyone can use our SitCov AV-testing framework and use it or build further on top of it. This paper aims to contribute to the domain of V&V and development of AVs, not only from a theoretical point of view, but also from the viewpoint of an open-source software contribution and releasing a flexible/effective tool for V&V and development of AVs.

## 1 Introduction

Autonomous vehicles (AVs) are no longer an idea of science fiction, they are being tested and even being used in some cities around the world. As seen in the past that with the adoption of new technologies, new kinds of hazards arise, similarly AVs have brought up a whole new concoction of hazards [1]. Since human lives are dependent when dealing with AVs, this makes AVs a safety-critical system. Hence, the safety of AVs cannot be taken callously.

**The Problem of AV Safety Assurance.** The safety standards applied to road vehicles currently, such as the ISO/PAS 21448:2019 [2] and ISO 26262 [3], do not translate well to AVs (SAE level 3 or above) [9], due to the fact that a human driver is required to take over in case of an emergency. An emerging standard UL-4600 [4] does not assume human drivers, but it only gives some guidelines to build a valid safety cases for autonomous systems, it is not at all prescriptive and can not be. Recently a new safety standard for automated vehicles has been published, the ISO 22737 [21], but it is quite limited. The ISO 22737 assumes low-speed automated driving, where routes are pre-defined within restricted operational design domains (ODDs).

In order to tackle this issue of safety of AVs, researchers have been employing various strategies, mostly using simulation softwares as a baseline since on-road testing of AVs is quite risky and costly. In [5] the authors attempt to model the distribution of disturbances over failures in a vehicle-to-vehicle (V2) interaction as a safety validation approach, this approach relies on selecting a particular set of disturbances to be injected. Moritz K. et al [6] proposes automatic critical scenario generation based on minimization of solution space, but this method relies quite a lot on discretization of solution space and constraining the behaviour of the vehicles. Greg C. et al [7] have proposed an agency-directed approach to test-suite generation for testing AVs, the scenario considered in their experiments is quite simple along with fidelity of simulation being quite low.

The papers mentioned above are quite recent but one thing lacking in these and most of the papers related to safety assurance of AVs is that the fidelity of simulators used for their experimentations, is quite low. Since the fidelity of the simulations is low, realistic camera images from the dash cam of the ego AVs can not be used since that functionality is not available in low fidelity simulators, and doing online-testing [8] of the perception system of AVs is really important while assuring the safety of the AV automated driving algorithm, since many AVs on roads these days rely on camera only (e.g., Tesla AVs). Along with no option of 3D rendering and live feed of dash-cam images being obtained by ego AVs in low fidelity simulators, the physics engine of low fidelity simulators is quite weak as well. The physics engine of such low fidelity simulators usually models just the basic equations of motions while neglecting some basic principles such a road friction, vehicles' tire friction etc.

**Exploration & Selection of AV Simulators.** With such considerations of limitations of AV simulators in mind, before moving on with the designs of our experiments and developing and testing our situation coverage-based (SitCov) AV-testing framework, we tried and tested a few AV simulators first before selecting CARLA, these are as follows: (1) MATLAB Automated Driving (AD) Toolbox [10]: This tool box has the upside of flexible scenario designing but the downside is that customizing experiments with AV Autonomous Emergency Braking (AEB) activated is not easily doable and there is no AV dash-cam option; (2) CarMaker [11]: This simulator has customizable scenarios with AEBs activated but it uses TCL script which is an outdated programming language and its interfacing with third party softwares is quite tedious and is not open-source; (3) CARLA [12]: Scenario customization is really easy, it has a high fidelity physics engine along with realistic 3D rendering and a long list of available sensors including the AV dash-cam. It uses Python language which is the goto programming language for researchers in various fields of Artificial Intelligence (AI). Also, CARLA is opensource which was one of the main aims of our research, to provide the public an effective AV-testing tool. Hence with all these upsides in mind, we selected CARLA as our simulator. Our code is publicly available here [20].

**AV Coverage-based Testing Methods.** Since AVs are a complex integration of systems of systems (SOSs), they face countless hazards due to the huge search space of inputs to the AVs. One methodology from software testing which is employed when testing a huge input space is called coverage criteria [13], which suits testing of AVs as well. Researchers have used the following coverage-based testing approaches for AVs recently [1]: (1) Scenario Coverage [15]; (2) Situation Coverage [14], [33]; (3) Requirements Coverage [14].

**Our Papers' Contribution.** In this paper we have developed a novel situation coverage-based (SitCov) AV-testing framework for the V&V and safety assurance of AVs and have used situation coverage as the coverage criterion for our automatic test suite generation, and we have come up with a unique/novel derivative of an ontology that we named situation hyperspace for our situation coverage-based situation generation, which we will elaborate in detail in the coming sections.

*Situation Coverage.* We define our situation coverage methodology as follows: *Situation coverage is a coverage criterion which takes into account the external and internal situations of the autonomous system (AV in our case), the automatic test suite generator (SitCov AV-testing framework in our case) uses the situation coverage metrics to know which situations have already covered and how many times each situation has been generated, based on this information the next batch of situations would be generated by the automatic test suite generator so that more of the situation space is covered and close to uniform distribution is achieved in case of repetitions of situations if all situations have been covered once.*

We would also like to define the term *situation* as *the initial configuration of the input space before the start of the simulation run (the temporal development of scenes).*

*Research Questions.* In order to quantify the situation space around our ego AV we have developed an ontology and used it to derive the situation hyperspace [16] which we will elaborate further in the next section. In this paper we investigate the following research questions (RQs): (1) RQ1: *How does situation coverage-based test suite generation perform against random generation from a viewpoint of revealing seeded faults?* (2) RQ2: *Does the situation coverage-based test suite generation provide any additional value in terms of the confidence in the safety metrics outputs of our SitCov AV-testing framework?*

We will look to answer these RQs in the subsequent sections. The rest of the paper is divided into the following sections. Section 2 presents the situation hyperspace that is used to quantify the situations around the ego AVs for our SitCov AV-testing framework. Section 3 breaks down the methodology and the development of the SitCov AV-testing framework in CARLA. Section 4 lays out the experimentation results and its in-depth analysis. Section 5 concludes the paper with some ideas for future work to build further on top of this proposed framework.

## 2 Situation Hyperspace

The literal meaning of hyperspace is "space of more than three dimensions/axis". By situation hyperspace [16], we refer to the multi-dimensional external world around the ego AV. This situation hyperspace has been constructed in a methodical way so that our SitCov AV-testing framework can systematically navigate through it to generate interesting and challenging situations for our ego AV using situation coverage-based generation to make sure good coverage of the situation hyperspace is executed by the SitCov AV-testing framework.

In order to come up with the ontology to derive the situation hyperspace for our framework, we have examined various AV world ontologies and ODDs. These include works by Krzysztof C. [17] in which the author has designed the operational world model (OWM) of the AV and AV ODD model presented by National Highway Traffic Safety Administration (NHTSA) [18]. We also studied the PhD thesis of Philippe N. [19] in which it is highlighted that road intersections are one of the highest risk areas among all road structures for road users with 30% of all road accidents occurring at intersections, with 14% of road accidents on intersections resulting in death.

After dissecting the literature on ontologies regarding AV ODDs we have come up with the situation hyperspace for our SitCov AV-testing framework as shown in Fig. 1 (left). In the block diagram of the situation hyperspace, we have the top layer as the situation hyperspace axis, it is further divided into environmental conditions axis and intersection axis. Other axis can be added to the situation hyperspace but right now since we were implementing this in our SitCov AV-testing framework in CARLA, we are using just two axes since we needed to keep the number low, of situation elements in the situation hyperspace and their combinations, which was practical when implementing it in our code. Nonetheless, these two axes are high priority axis as per literature that is another reason we have added and tested them first. The SitCov AV-testing framework selects the situation elements from both environmental conditions and intersection axes and combines them to generate the discrete situation where the AV simulation actually runs in CARLA.

**The Environmental Conditions Axis.** For the environmental conditions axis, we further subdivided it into the following situation elements, after studying a variety of AV ODD and world ontologies [17, 18]: (1) Friction (road friction); (2) Fog Density; (3) Precipitation; (4) Precipitation Deposits; (5) Cloudiness; (6) Wind Intensity; (7) Wetness; (8) Fog Distance. These situation elements of the environmental conditions axis can be seen in Fig. 2, each element has been discretized into 6 discrete bins and our SitCov AV-testing framework chooses from those.

**The Intersection Axis.** In the intersection axis, T intersection has been chosen as the axis to do AV-testing on since it has one of the highest accident rates [19]. We have also highlighted other axis types that could be added to the intersection axis, such as 4-legged intersection, offset 4-legged intersection, circular 4-legged intersection, different skew angles of intersection legs in all types. But for the sake of simplicity and implementation we have chosen T-intersection as the only element of the intersection axis.

In our SitCov AV-testing framework, we are essentially doing pairwise testing of our ego AV with an adversarial other vehicle which we will now refer to as other vehicle (OV). The reason for selecting pairwise encounters of the ego AV with the OV is that V2V accident rates on intersections are way above all other types of accidents on intersections as mentioned here [19]. The pairwise testing between the ego AV and OB is done on a T-intersection in different environmental conditions selected from the environmental conditions axis by our framework, and in different intersection configurations i.e., different start and goal locations for the ego AV and OV, selected from the intersection axis, as seen in Table-1. For the different intersection configurations, we have come up with collision-points $c_1$, $c_2$, $c_3$, $c_4$ as seen in the center of the T-intersection in Fig. 1(right), these are the closest point of approach (CPA) for our ego AV and OV for their pairwise interaction during each simulation run. This CPA approach has been inspired from the work of [22]. The four collision-points we have come up with are one of the highest risk areas on an intersection and are similar to the conflict-points on intersections mentioned in [17]. Hence, testing our ego AV on one of the most dangerous road structures (intersection) on the highest risk areas (intersection conflict-points) would be an effective way to stress the ego AV and test it thoroughly, which our SitCov AV-testing framework is doing.

For the pairwise AV-testing on the T-intersection, we will consider right-hand traffic, the main reason being that many road ontologies that we have followed, used right-hand traffic. There will be three possible start and goal locations for the vehicles (AV on the T-intersection. As we would only consider the vehicles to be moving inwards towards the T-intersection after spawning at their start locations, we are considering cross-collisions [23] at the moment, we will incorporate diverging and merging type collisions [23] in the later stages also. The SitCov AV-testing framework generates the OV such that it comes in conflict with our ego AV at the CPA i.e., one of the collision-points ($c_1$, $c_2$, $c_3$, $c_4$).

**Notations For Intersection Pairwise Conflict-Point Interactions.** Notations for the intersection pairwise conflict-point interactions between ego AV and the OV (as seen in Table-1) are elaborated as follows: (1) *L, R,* and *B*, define left, right and base legs of the T-intersection; (2) *S* implies start location variable; (3) $S_x$, *x* subscript defines the type of vehicle. It can either be AV or OV in the pairwise encounter; (4) $S_{xy}$, *y* subscript defines the location of the start point *S* of the vehicle; (5) *G* implies goal location variable; (6) $G^z$, *z* superscript indicates the final destination/goal location of the vehicle; (7) The multiply operator "x", is the intersection pairwise conflict-point interactions operator between the variables as seen in the "Conflict Point Interaction" column of Table-1, the notations in that column essentially tells what were the start/goals locations of the ego AV, the OV and the particular conflict-point where the ego AV and OV met during a simulation run by the SitCov AV-testing framework.

Intersection pairwise conflict-point interactions when ego AV starts at the base, have been shown in Fig. 1(right) along with their notations and can be compared with Table-1. The complex notations shown as conflict-point interaction column in Table-1 have been simplified in the last column of the Table-1, i.e., intersection situation labels, and these simplified labels will be used to represent these complex conflict-point interactions between our ego AV and the OV. These intersection situation labels are the situation elements of our intersection axis of the situation hyperspace. We encourage the readers to see how these complex conflict-point interactions of the intersection axis have been modelled in our code [20] using dictionaries and lists in Python language.

The next section elaborates how does the SitCov AV-testing framework selects the situation elements from environmental conditions and intersection axis of the situation hyperspace, to generate the discrete situation in which the V2V interaction between the ego AV and OV takes place in the CARLA simulator. The next section also highlights key elements in the development of the SitCov AV-testing framework in CARLA.

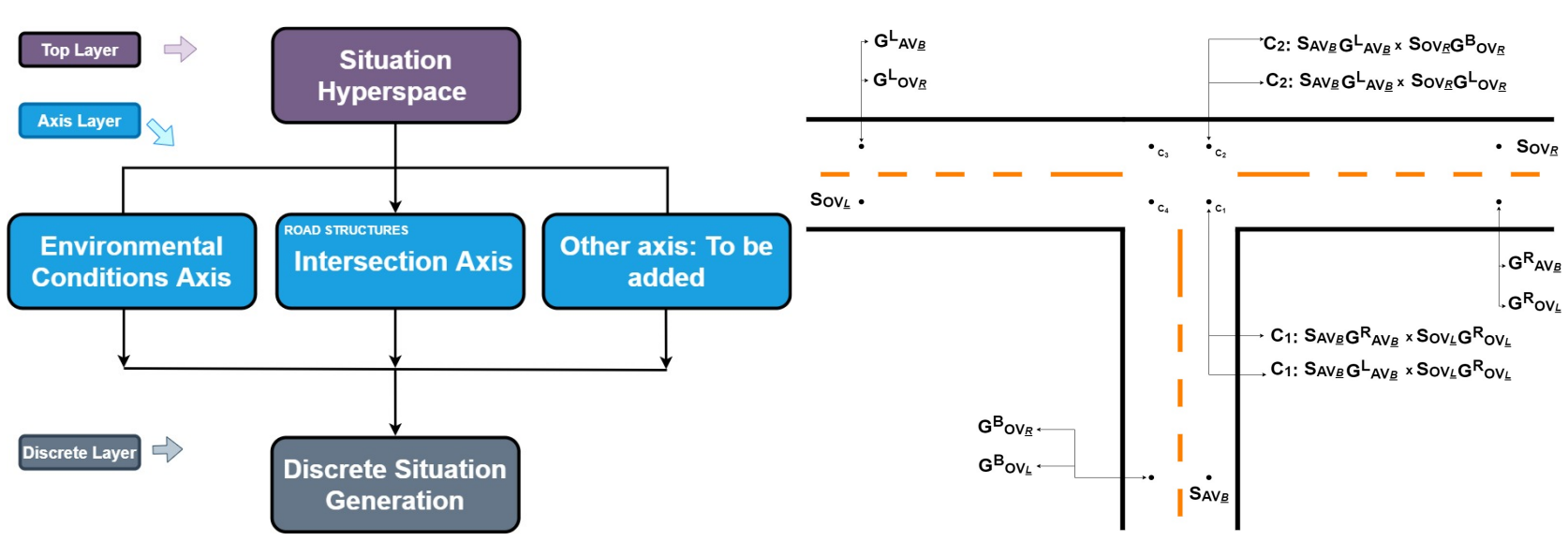

**Fig. 1.** Situation hyperspace block diagram (left). T-intersection with notations (right).

**Table 1. Intersection Situations & Notations**

| Sr | Start loc of AV | Goal loc of AV | Conflict Point Interaction | Intersection Situation Label |
|---|---|---|---|---|
| 1 | $S_{AV\underline{B}}$ | $G^{R}_{AV\underline{B}}$ | $C_1$: $S_{AV\underline{B}}\ G^{R}_{AV\underline{B}} \times S_{OV\underline{L}}\ G^{R}_{OV\underline{L}}$ | IntSit-8 |
| | | $G^{L}_{AV\underline{B}}$ | $C_1$: $S_{AV\underline{B}}\ G^{L}_{AV\underline{B}} \times S_{OV\underline{L}}\ G^{R}_{OV\underline{L}}$ | IntSit-5 |
| | | | $C_2$: $S_{AV\underline{B}}\ G^{L}_{AV\underline{B}} \times S_{OV\underline{R}}\ G^{L}_{OV\underline{R}}$ | IntSit-6 |
| | | | $C_2$: $S_{AV\underline{B}}\ G^{L}_{AV\underline{B}} \times S_{OV\underline{R}}\ G^{B}_{OV\underline{R}}$ | IntSit-7 |
| 2 | $S_{AV\underline{R}}$ | $G^{B}_{AV\underline{R}}$ | $C_2$: $S_{AV\underline{R}}\ G^{B}_{AV\underline{R}} \times S_{OV\underline{B}}\ G^{L}_{OV\underline{B}}$ | IntSit-10 |
| | | | $C_4$: $S_{AV\underline{R}}\ G^{B}_{AV\underline{R}} \times S_{OV\underline{L}}\ G^{R}_{OV\underline{L}}$ | IntSit-11 |
| | | | $C_4$: $S_{AV\underline{R}}\ G^{B}_{AV\underline{R}} \times S_{OV\underline{L}}\ G^{B}_{OV\underline{L}}$ | IntSit-12 |
| | | $G^{L}_{AV\underline{R}}$ | $C_2$: $S_{AV\underline{R}}\ G^{L}_{AV\underline{R}} \times S_{OV\underline{B}}\ G^{L}_{OV\underline{B}}$ | IntSit-9 |
| 3 | $S_{AV\underline{L}}$ | $G^{B}_{AV\underline{L}}$ | $C_4$: $S_{AV\underline{L}}\ G^{B}_{AV\underline{L}} \times S_{OV\underline{R}}\ G^{B}_{OV\underline{R}}$ | IntSit-1 |
| | | $G^{R}_{AV\underline{L}}$ | $C_4$: $S_{AV\underline{L}}\ G^{R}_{AV\underline{L}} \times S_{OV\underline{R}}\ G^{B}_{OV\underline{R}}$ | IntSit-2 |
| | | | $C_1$: $S_{AV\underline{L}}\ G^{R}_{AV\underline{L}} \times S_{OV\underline{B}}\ G^{R}_{OV\underline{B}}$ | IntSit-3 |
| | | | $C_1$: $S_{AV\underline{L}}\ G^{R}_{AV\underline{L}} \times S_{OV\underline{B}}\ G^{L}_{OV\underline{B}}$ | IntSit-4 |

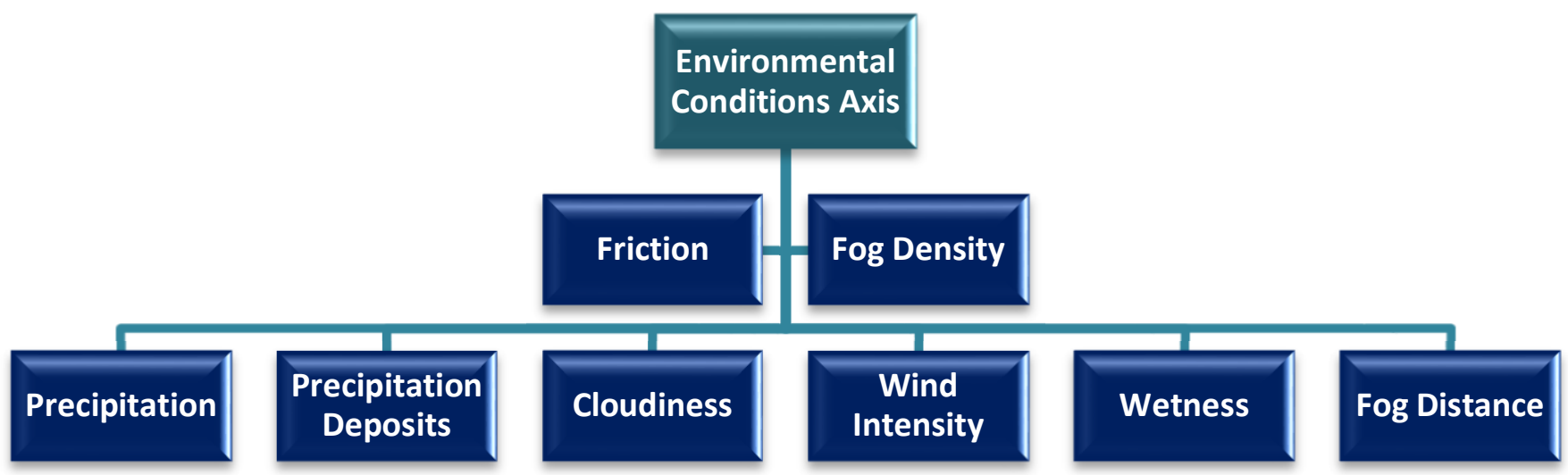

**Fig. 2.** Environmental conditions axis subdivision elements.

## 3 The SitCov AV-Testing Framework

This section lays out the development of the proposed SitCov AV-testing framework from top to bottom, from the theory to its implementation in CARLA. Below we start with explaining the theoretical mechanics of our SitCov AV-testing framework along with its connection with test adequacy criteria from software testing [25].

### 3.1 Situation Coverage-based AV-testing Test Suite Generation Methodology

In this subsection we will briefly explain the methodology behind the situation coverage-based situation generation for AV-testing by our SitCov AV-testing framework.

**Test Adequacy Criteria for AV-Testing.** The inspiration of the SitCov AV-testing Framework has been taken from the concept of test adequacy criterion from software testing [25], which essentially states that test adequacy criterion provides a stopping rule and/or a measure of test quality. A test adequacy criterion highlights the testing requirements and the test suits needed to satisfy those requirements, and it determines the observations required during the process of testing.

In this paper, we have made a connection between this concept of test adequacy criterion from software testing with AV testing, by applying it in our SitCov AV-testing framework as seen in Fig. 3 (right). Our SitCov AV-testing framework utilizes situation hyperspace for situation coverage metrics along with fault injection. This will provide us with reliable test adequacy criteria-based metrics. Fig. 3 (right) shows the test adequacy criteria highlighted in blue being employed by our SitCov AV-testing framework, further detailed definitions of these test adequacy criteria along with their connection with our framework are out of the scope of this paper, the test adequacy criteria definitions can be found here though [25].

**Handling the AV Nominal vs Functional Safety Problem.** The safety assurance of AVs is particularly a challenging task due to AVs being a cyber-physical, we not only have to make sure that the functional safety of the AV is addressed i.e., software/hardware of the AV is bug-free and all the functions are being executed correctly, but also that, we have to make sure that we have addressed the nominal safety [26] aspect of the AV as well i.e., making sure the AV is making safe and logical decisions, assuming that the software/hardware of the AV are operating error-free.

To address both the issues of AV functional and nominal safety, we will use the fundamentals provided by software testing as described above in conjunction with agent-based simulation in CARLA.

**Automatic Test Suite Generation of our SitCov AV-testing Framework.** The situation coverage-based generation of the discrete situations from the situation hyperspace is done by counting how many times each bin of the situation elements of each of the environmental conditions and intersection axis, has been generated. Then the counts (number of times that bin of the situation element was used for a discrete situation generation simulation run) of each bin of a particular situation element of an axis of the situation hyperspace, let's say the precipitation element from the environmental conditions axis, or the intersection situation labels from the intersection axis as mentioned in Table-1, are fed into a SoftMax function to generate a normalized probability distribution over those bins, then those probability values are inverted and normalized and act as weights to a weighted random number generator, higher the weight, higher the probability of that bin to be selected. In simple terms, those bins of a situation element, let's say the precipitation element of the environmental conditions axis or the intersection situation labels from the intersection axis, that have been generated fewer times than the other bins (for discrete situation generation), have the highest probability to be selected by the SitCov AV-testing framework for the discrete situation generation in the next simulation run. This way we aren't setting hard rules that all situations have to be generated exact equal number of times, we are leaving some room for exploration along with exploitation by using weighted random number generation while still making sure that with a very high probability good coverage of the situation hyperspace is being achieved. This whole process of situation coverage-based generation from the situation hyperspace can be seen in our code here at [20], this process can also be seen in the block diagram of the SitCov AV-testing framework in Fig. 3 (left), along with other processes that are occurring such as keeping track of collisions happening between ego AV and OV, and increasing the counter of those situation element bins that were used to generate the last discrete situation for the simulation run, etc.

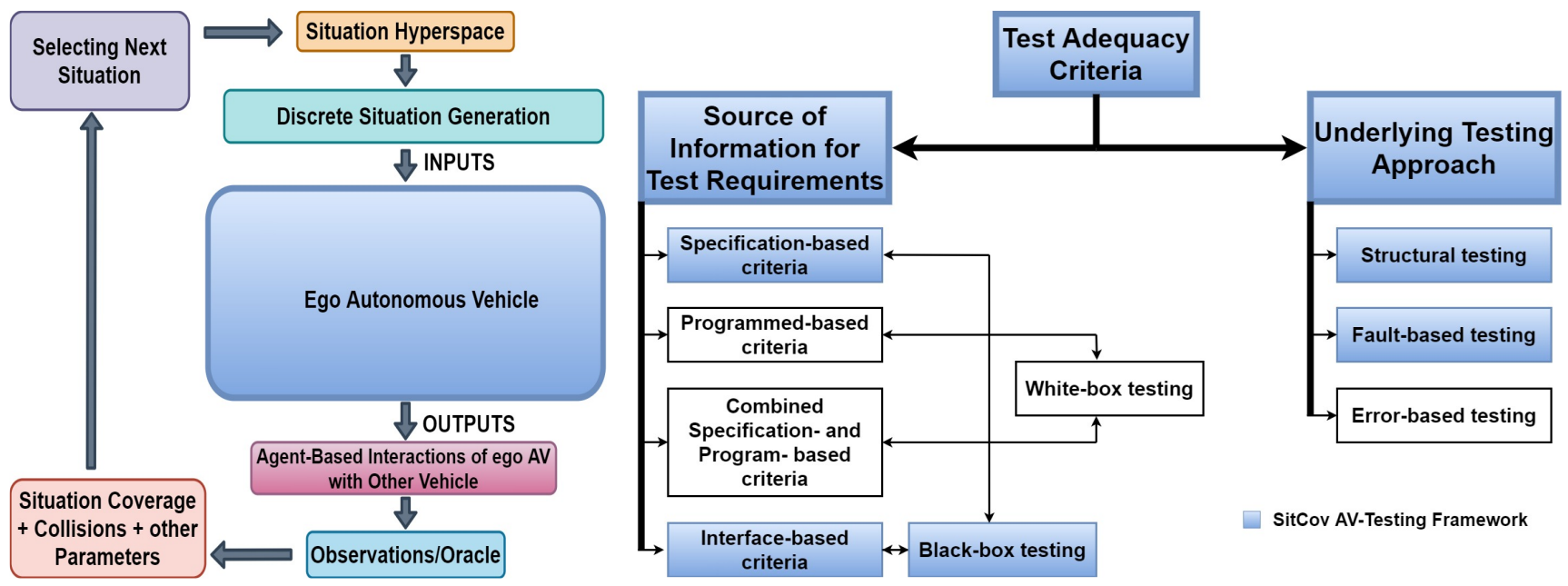

**Fig. 3.** SitCov AV-testing framework block diagram (left). Test adequacy criteria used (right).

**The AV-IP Problem & Our Solution.** A really important but seldom talked about problem for the practical V&V and safety assurance of AVs is the IP (Intellectual Property) problem. As top tech (Waymo, Apple, etc.) and car manufacturing (Tesla, BMW, etc.) companies compete with each other in the AV domain, with projected trillions of dollars of revenue on the line, these companies are investing millions and billions of dollars in the AV industry, and to say the least, they would not be very pleased to share their entire AV IPs with third party safety experts for the sole purpose of V&V and safety assurance of AVs.

Keeping in view, this IP problem of AVs, we have designed our SitCov AV-testing framework to treat AVs as a black-box or in some instances as a grey-box, which will enable our research to be industry ready and hopefully contribute directly to saving lives in the process. The users of our SitCov AV-testing framework will just have to insert their AV autonomous driving stack as a black-box and run the SitCov AV-testing framework, the framework will output situation coverage metrics of all situation elements of the situation hyperspace and those situations that the AV autonomous driving stack cannot deal with will be highlighted along with the percentages of the situation covered in the situation hyperspace and their repetitions. This process will be further elaborated in the experiments results and analysis section.

### 3.2 Developing The SitCov AV-Testing Framework in CARLA

This subsection presents the main softwares used for the development and experimentation of the SitCov AV-testing framework. Following were the baseline software packages used: (1) CARLA [12]; (2) Scenario Runner [27]; (3) Tensorflow Object Detection API [28]. All of these software packages are opensource and Python is the programming language used to run them. A few screenshots of the development of the SitCov AV-testing framework can be seen in Fig. 4. Further elaboration of development of the framework is below.

**Using CARLA & Scenario Runner for Simulations.** As stated in the previous sections that CARLA was chosen due to its upsides as compared to other AV simulators that includes it being opensource, regularly updated with previous versions maintained on GitHub website, 3D rendering and AV dashcam video feed are available along with other sensors. We have used Scenario Runner [27] on top of CARLA for the development of our SitCov AV-testing framework. Scenario Runner is just an API built on top of CARLA software and it is used to generate certain scenarios. The intersection scenarios generation from Scenario Runner was of our interest and we edited that intersection class to run our AV-testing test suites on a T-intersection by generating situation coverage-based situations from the situation hyperspace that we designed in Python using dictionaries and lists [20].

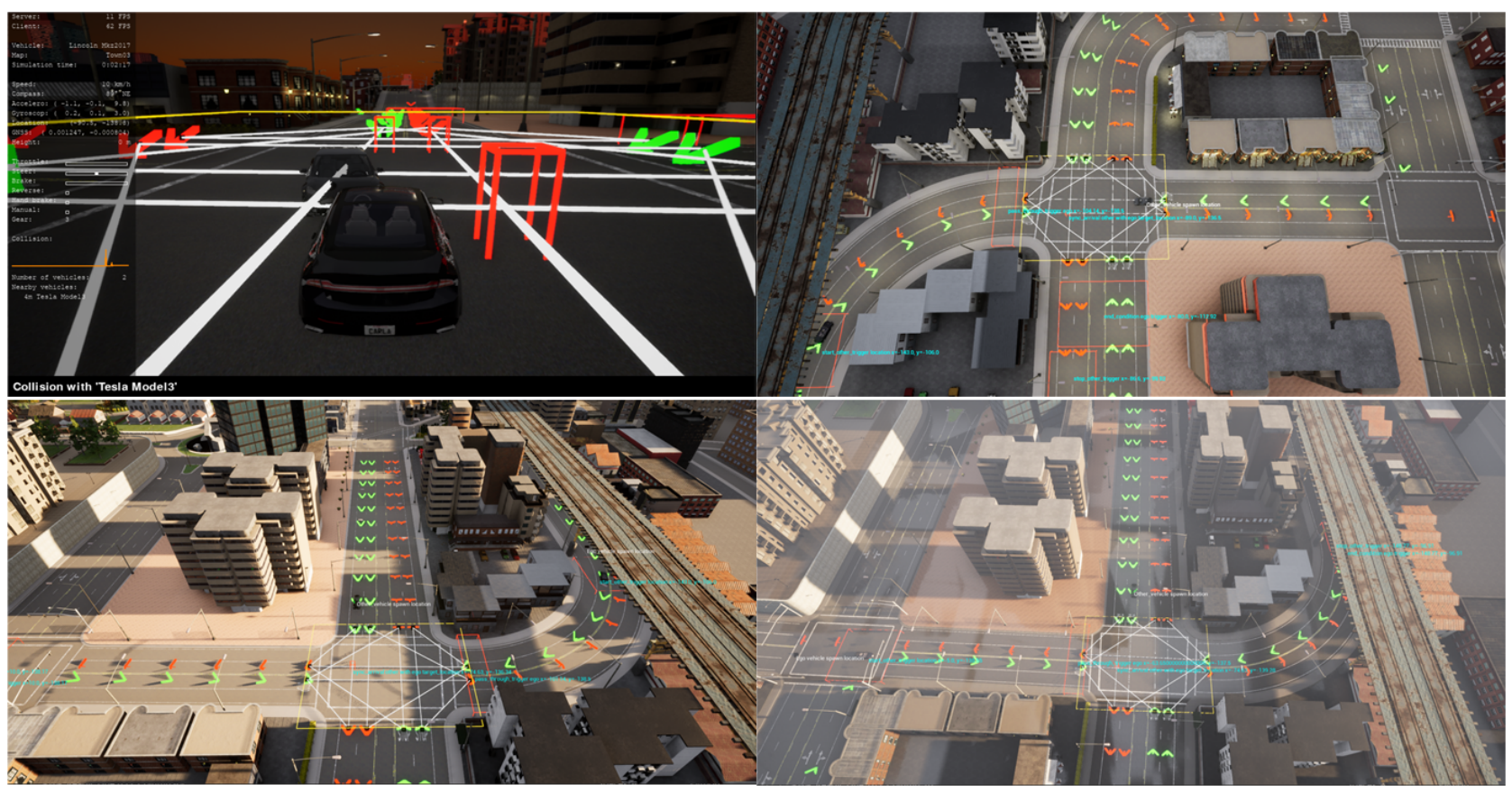

**Fig. 4.** Development of our SitCov AV-testing framework in CARLA and Scenario Runner. The tools provided in CARLA such as waypoints highlighting and their interconnection with each other and other road structures such as intersections etc., were really helpful in speedy development of our SitCov AV-testing framework.

**Developing Automated Driving Algorithm for Ego AV using Tensorflow Object Detection API in CARLA.** In order to evaluate our SitCov AV-testing framework we needed an autonomous driving (AD) algorithm for the ego AV in our simulation runs, and we also need an algorithm to drive the OV as well since our SitCov AV-testing framework ensures that the ego AV and OV meeting in the intersection at the collision-points shown in the previous section, under different situations generated from the situation hyperspace using situation coverage-based generation. So, for the ego AV autonomous driving, ideally, we would have liked to get that algorithm from an OEM such as Tesla, then our SitCov AV-testing framework would have been properly evaluated, since we would have had high confidence that there was a very low probability of the ego AV crashing in the OV due to faults in the autonomous driving algorithm. Even though our SitCov AV-testing framework would have treated the ego AV autonomous driving algorithm as a black-box, we still wouldn't have been able to get such an algorithm from an OEM easily. So, we had to develop it on our own autonomous driving algorithm for the ego AV.

*Types of Autonomous Driving Algorithms.* We had the option to choose from 3 overarching pipeline of autonomous driving as mentioned in [12]: (1) Modular pipeline; (2) Imitation learning; (3) Reinforcement learning. Imitation learning and reinforcement learning pipeline of autonomous driving isn't very effective when it comes to rule-based driving. Imitation learning does perform a lot better than reinforcement learning in rule-based driving [12] but modular pipeline is the best option for rule-based driving and it is what actual driving is like in the real world as well, traffic-rules-based. Hence, we chose modular pipeline for our ego AV autonomous driving.

*Our Modular Autonomous Driving Pipeline.* Our modular autonomous driving pipeline has following main stages: (1) Perception; (2) Local Planner; (3) Continuous Controller. The local planner stage was already implemented in CARLA examples, which used A-star algorithm [29] to compute the optimal path from one point of the map to the other using the roads and following traffic rules. We used this directly and our assumption here is that our ego AV has an ideal GPS sensor which tells it the exact location because the local planner we are using is accessing the ground truth for accessing the location of the location of the ego vehicle which is the ego AV or OV, since we are using this local planner in both.

The continuous controller stage is a PID controller for lateral and longitudinal control of the ego vehicle, and this was also already implemented in CARLA that we used directly for the ego AV and OV controllers.

*The Perception Stage (Using Deep Neural Network).* The ego AV uses one more stage for its autonomous driving that the OV does not, it is the perception stage. Just like Tesla AVs, we are using only camera as the perception sensor for our ego AV. We are using a very deep Single-Shot multi-box Detection (SSD) Convolutional Neural Network (CNN) pre-trained on 350,000 images of the MS COCO dataset, the SSD mobilenet [28, 30, 31], to detect the incoming OV and locate its position in the dash-cam feed (images) of the ego AV and to apply emergency brakes (AEBs) if the OV is too close. We are using the Tensorflow Object Detection API to implement this pre-trained SDD mobilenet CNN.

*Parameters of Object Detection.* The perception stage of our ego AV has three main parameters that can be set for efficient object detection using the pre-trained SDD mobilenet CNN that is taking images from the dash-cam of the ego AV in CARLA. Following are the three parameters:

1. **Threshold of distance for object detection:** To detect the distance of the OV from the image we are using a computer vision technique [32] to check how much percentage of the image feed is the detected OV taking up, the closer is the OV to the ego AV, the more percentage of the image from the ego AV would contain the OV and at a certain threshold value, our ego AV will apply the AEBs to avoid collision.

We will refer to this threshold value as the "threshold of distance for object detection" in the later sections.

2. **Probability of detection for object detection threshold:** This is the threshold value for probability output from the SDD mobilenet CNN, which tells us the probability if there is a car detected in the images received from the dash-cam of our ego AV in CARLA. Since we are doing V2V pairwise testing with an OV, we only need to care about the probability of detecting a car in the received images, though we could also check for detection of buses, trucks, people, animals, etc., we recommend including other dynamic agents in future work. Setting this parameter for probability of detection is really important, i.e., the minimum probability value that our ego AV would consider as a detection of an OV in the image. Since too high a value of this parameter would result in missing detection and collisions happening in extreme weathers such as heavy rain and fog etc., where probability of detection of OV would be lower even if it was in front of our ego AV.
3. **Centering limits parameters:** These parameters look at if the OV detected by our SDD mobilenet CNN is close to the center of the received image. If the detected OV is too far away from the center, even if the OV is really close, AEBs are not activated,

So, this makes up the autonomous driving pipeline of our ego AV, provides a useable driving pipeline to test the main focus of this paper, the SitCov AV-testing framework and its implementation in CARLA. A bit of effort was also put into implementing the autonomous driving modular pipeline in CARLA as elaborated above and we would like to recommend the readers to have a look at its implementation in the code if they're looking to get a working implementation of autonomous driving [20]. Next, we will show the screenshots of a few experiments, i.e., the discrete situations generated by our SitCov AV-testing framework, just so that the readers get a gist of how these experiments are happening practically in CARLA software with the test suites generated by our SitCov AV-testing framework.

**Running Test Suites on the SitCov AV-Testing Framework.** In Fig. 5 we can see experiment #1 generated by the SitCov AV-testing framework. The temporal progression of the simulation run starts from top left image to the top right, then the bottom left and the last timestamp is the bottom right picture in Fig. 5. The SitCov AV-testing framework selected a sunny day with clear weather and generated the ego AV on the left side of the intersection and the OV was generated on the top side of the intersection as seen in the first picture of the temporal sequence of experiment #1, all these situations were selected from the situation hyperspace using situation coverage-criterion by our framework. In all the pictures we can see another window opened up in the top right corner of each picture, this window is showing the images received by the dash-cam of the ego AV and it is also showing the object detection probability values and detection boxes if an OV is detected. As seen in the third picture of the temporal sequence, the ego AV has detected the OV really close to it and close to its centers as well (the path of progression of the AV) and the probability of detection is above the threshold value we have set, hence AEBs are activated and the ego AV comes to a full stop and a collision is avoided and the experiment is a success. This OV can be seen going away in the last picture of the temporal sequence.

Fig. 6 shows experiment #2, in which the SitCov AV-testing framework has generated a heavy rain and a foggy discrete situation from the situation hyperspace. The OV is coming in from the top of the intersection and the ego AV is coming in from the right side of the intersection. Due to the heavy rain and fog, the ego AV was not able to detected the OV in time for AEB and a collision happened between the two. This accident was avoidable if the ego AV had detected the OV just a few moments earlier.

Fig. 7 show experiment #3, in which there is just a little bit of rain but heavy fog and a lot of precipitation deposits on the road, selected by our SitCov AV-testing framework. The OV is coming in the intersection from the left side and the ego AV is coming in from the right side and the OV suddenly turns towards the ego AV and collides. There was much the ego AV could do as both vehicles were following the traffic rules before the OV suddenly turned towards the ego AV and collided. We will refer to such accidents/failures as unavoidable accidents/failures.

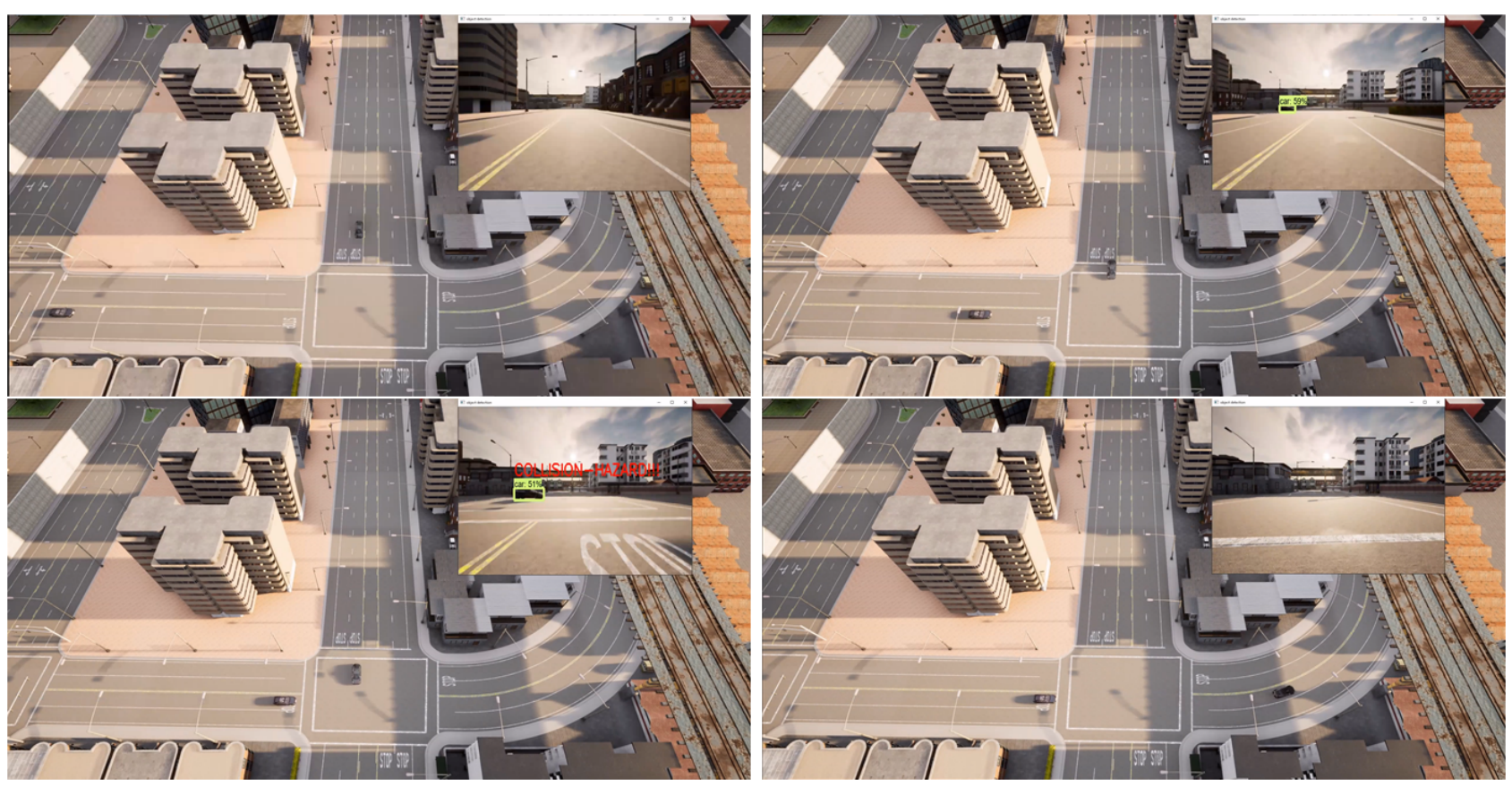

**Fig. 5.** Experiment #1 Sunny Day Accident Avoided: Success.

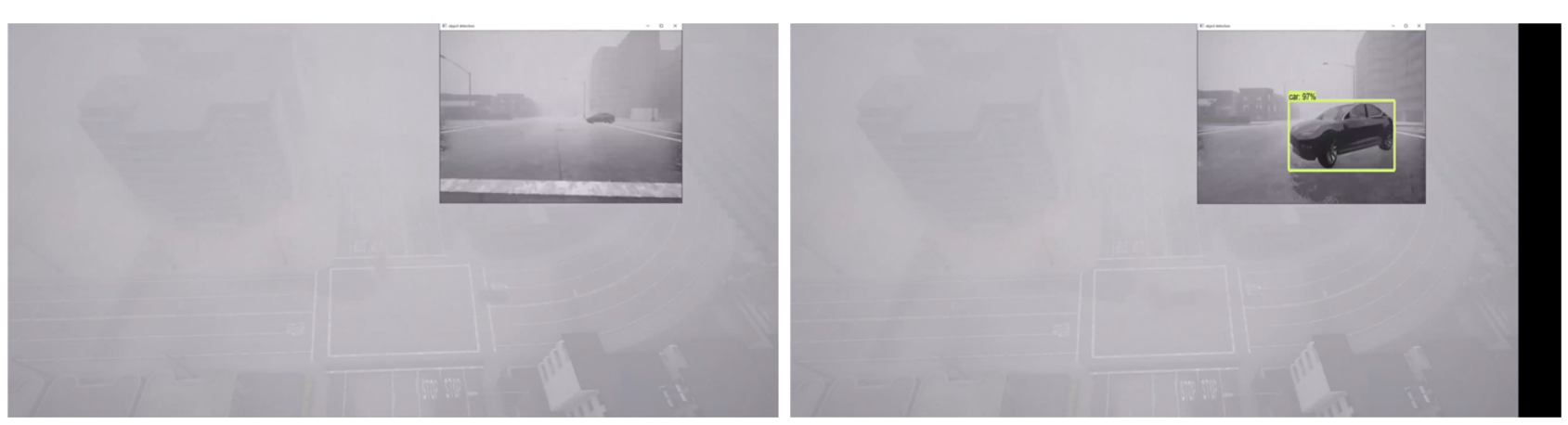

**Fig. 6.** Experiment #2 Heavy Rainy & Foggy Day & Avoidable Accident Collision: Fail.

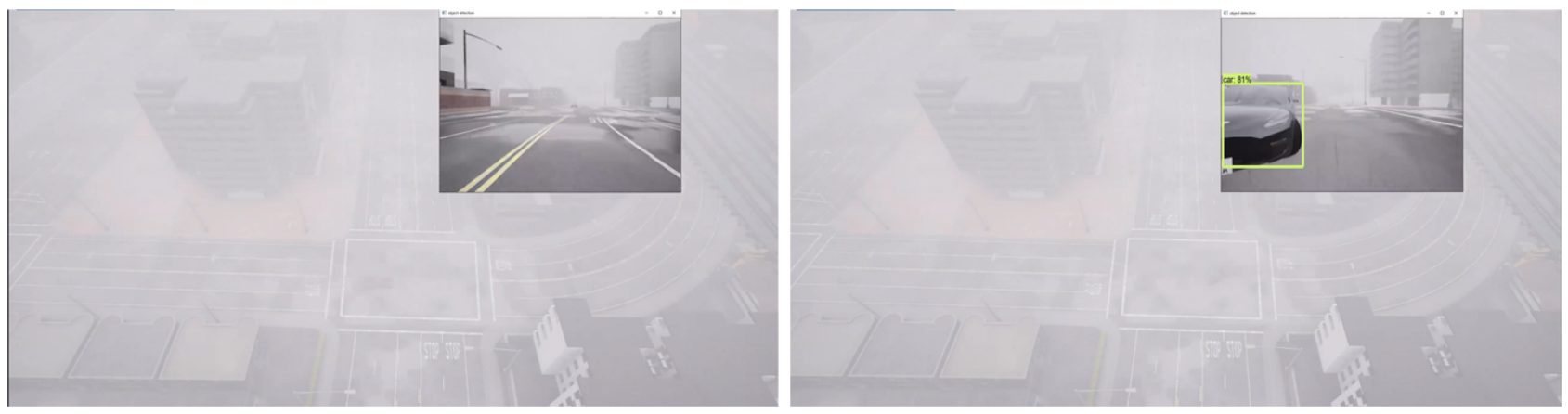

**Fig. 7.** Experiment #3 Rainy Foggy Day & Unavoidable Accident Collision: Fail.

## 4 Experimentation Results & Analysis

We provide the details of our experimentation setup and a detailed analysis and evaluation of our SitCov AV-testing framework in this section. We have carried out experiments to evaluate our SitCov AV-testing framework and our goal is to answer our research question RQ1 and RQ2, that how well does our SitCov AV-testing framework perform compared to random situation generation, and what additional useful information does our SitCov AV-testing framework gives us that random situation generation cannot.

For the experimentations we used a core-i7 PC with a 4GB GPU and 32 GB RAM. We ran experiments in sets of 20, and repeated it 5 times with a different starting random seed, to make it a total of 100 experiments. Each set of 20 experiments took almost 30 minutes to run, so one set of 100 experiments took about 2.5 hours.

**Evaluation Metrics and Seeding Faults.** We use seeded faults to evaluate our SitCov AV-testing frame. The process starts with us seeding a few faults in the ego AV software, and when the ego AV crashes in the OV, we say that the *seeded fault* has been *triggered*. But there is a caveat, as we mentioned in the previous section of our autonomous driving algorithm development, we have used a fairly simple algorithm and just one RGB-camera sensor on our ego AV, hence there will be a lot of non-seeded background fault causing failures (crashes of ego AV in OV) and apart from that there will be some unavoidable accident situations as shown in Fig. 7 that when the OV suddenly turns towards the ego AV.

So, we will denote crashing of ego AV into the OV as a failure, whether it is due to a seeded fault being triggered or it is due to non-seeded background faults or simply unavoidable failures as mentioned above. Then we will first conduct experimentations without seeding any faults and note how many failures we get per 100 experiments which we will call as *general failures* (which include both non-seeded background faults and unavoidable accident situations), then we will seeded faults into the same 100 experiments (with the same random seeds, so that the discrete situations generated are the same as what they were for no seeded faults experimentations), and the extra failures counts exceeding the general failure counts will be labelled as those seeded faults being triggered.

**Seeded Faults.** Following are the three types of faults that have been seeded in our ego AV software, we have targeted parameters of object detection, detailed in the previous section, for seeding our software faults:

1. **Fault #1:** Setting *Probability of Object Detection Threshold* really high, i.e., to 0.95. Our ego AV will detect the OV in front of it only if the probability of its detected outputted by the deep convolutional neural network is 0.95.
2. **Fault #2:** Setting the *Centering limits parameters* too rigid, i.e., our ego AV will only label the OV in front of it as a hazard, even if it is dangerously close to it, if the OV is in the exact center of the image received by the ego AV dash-cam.
3. **Fault #3:** Setting the *Threshold of distance for object detection* to a really low value, i.e., our ego AV will only consider the OV as a hazard if it is extremely close to it because we have lowered the threshold of distance for object detection parameter.

### 4.1 Intersection Situation Generation Results

In this subsection we are highlighting the performance of our SitCov AV-testing framework vs random situation generation w.r.t the intersection situations being generated from the intersection situation axis as mentioned in Table-1. These experiments will explore which intersection situations are dangerous for our ego AV, having the highest failure rates.

**No Faults Seeded Intersection Situations Experiments.** Our first set of experiments compares intersection situation generation results of our SitCov AV-testing framework vs random generation, when no faults have been seeded.

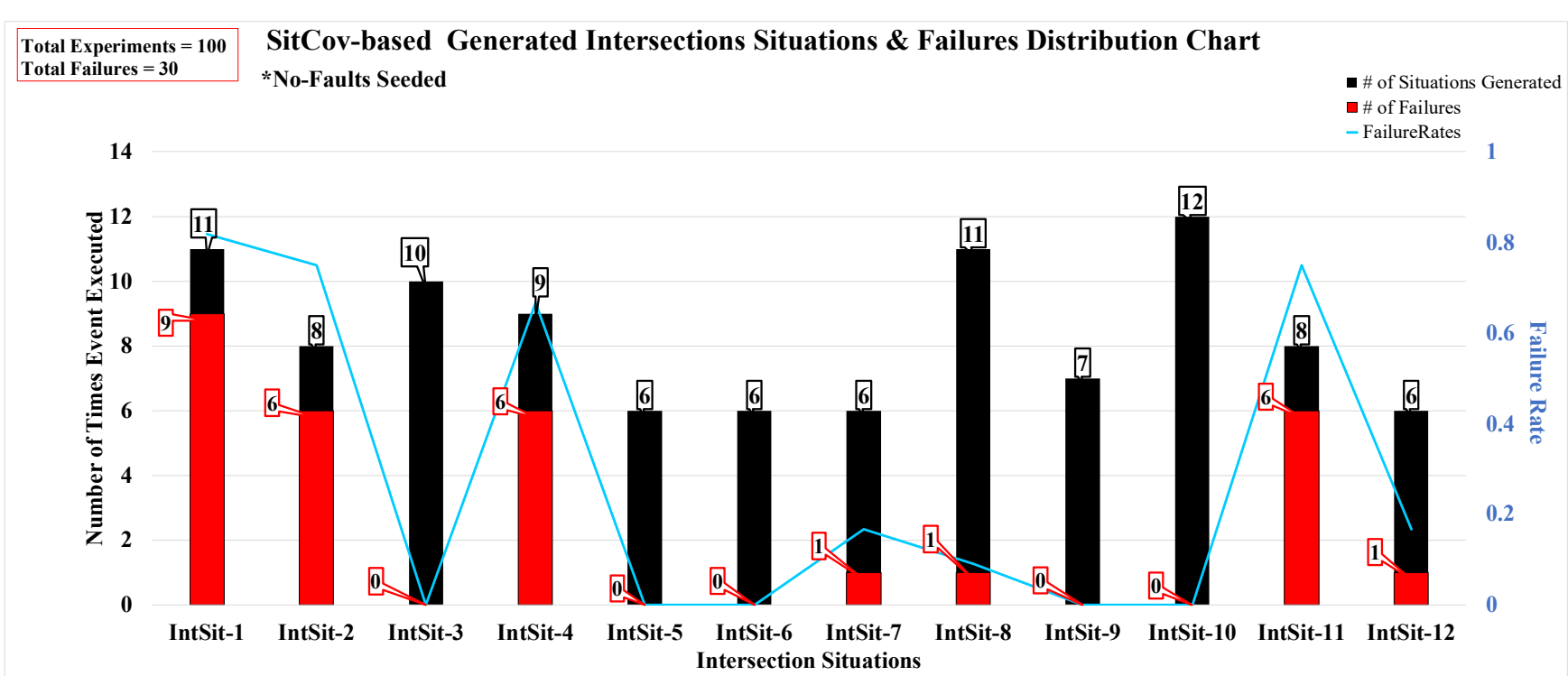

**Fig. 8.** SitCov-based intersection situation generation & failure distributions, no faults seeded.

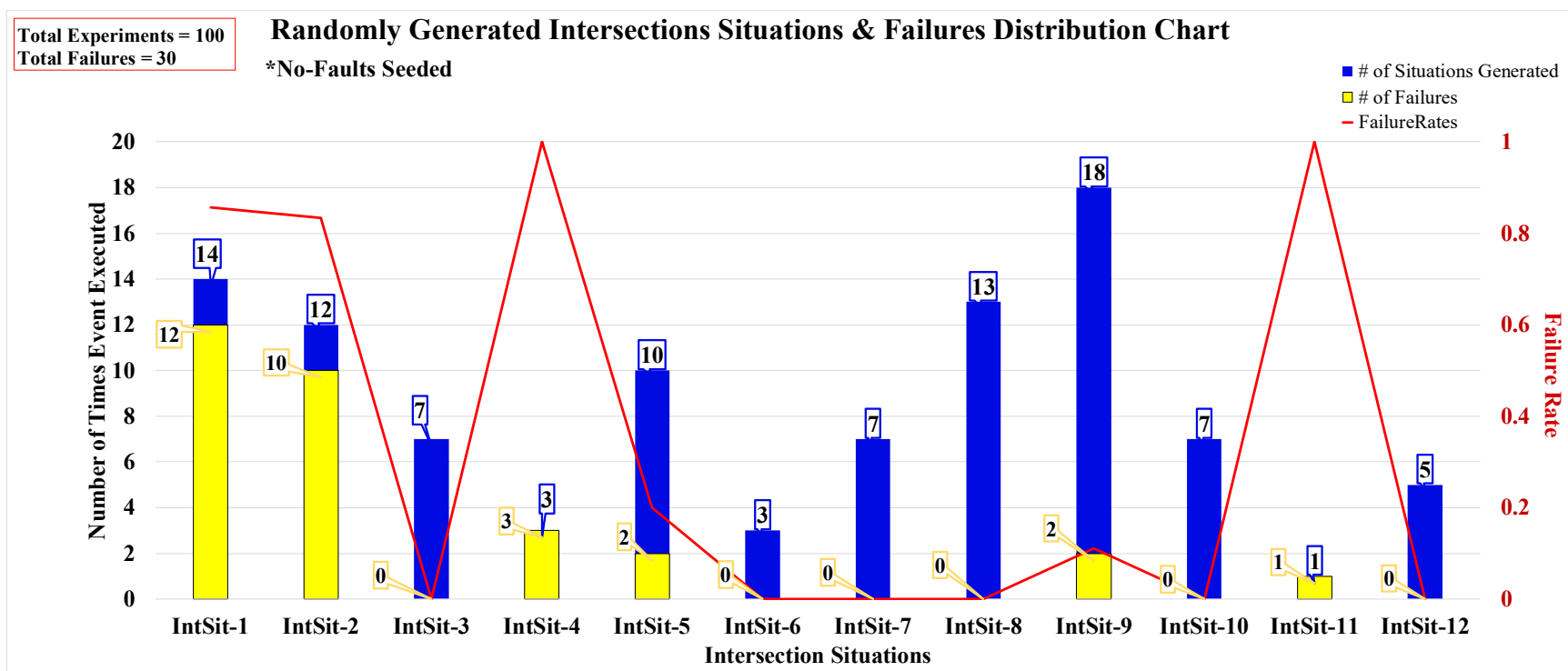

**Fig. 9.** Random intersection situation generation & failure distributions, no faults seeded.

In Figs. 8, 9 the labels mentioned on the x-axis are the intersection situation labels that have been shown in detail in Table-1, which mention particular intersection situations as derived from the notations used and explained in the previous sections. The labels will be used in the coming figures as well.

As seen in Fig. 9, random intersection situation generation has quite erratic failure rates values vs the situation coverage-based (SitCov-based) intersection situation generation from our SitCov AV-testing framework in Fig. 8, even though their total number of failures are equal. This is due to the uneven distribution of situations produces by the random generation, which is why we can see for some intersection situations the failure rate is 100% for random generation in Fig. 9, where as this is not the case for situation coverage-based intersection situation generation in Fig. 8, because each intersection situation was tested somewhat evenly and hence we can have more confidence in the failure rates of the intersection situations provided by our SitCov AV-testing framework and then we can in-turn look at the ego AV autonomous driving algorithm and see why does it have high failure rates in these particular intersection situations and how can the autonomous driving be improved.

**Faults Seeded Intersection Situations Experiments.** Fig. 10 compares the results of SitCov AV-testing framework with random generation. On the left colored in red and black is our framework's results and on the right in yellow and blue is the random generation results. The top row results are when fault #1 is seeded, the middle row results are when fault#2 is seeded and the bottom row results are when fault #3 is seeded.

Again, we can see that the distribution of intersection situation generation of our SitCov AV-testing framework is quite uniform as compared to the random generation, hence the failure rates are more reliable. Hence, we will analyze the results of our SitCov AV-testing framework from Fig. 10, below.

As seen in Fig. 10, fault #1 is triggered the most, which tells us that the parameter *Probability of Object Detection Threshold*, that we tinkered as a part of our fault #1 seeding, is extremely important for autonomous driving.

After fault #1, fault #2 is triggered the most (Fig. 10), which refers to the *Centering limits parameters*, which detects the OV as a hazard only when it is within a certain distance from the center of the line of sight of ego AV, i.e., from the center of the image taken by the dash-cam on the ego AV. This means that the developers need to be wary of not only the vehicles coming in the line of sight of the ego AV, but also vehicles further out in the field of view (FOV) of the ego AV.

Fault #3 is triggered the least (Fig. 10) but its failure rate is still really high as compared to when no faults were seeded (Fig. 8), which highlights the importance of the parameter which was tinkered as a part of fault #3, i.e., the *Threshold of distance for object detection*. This tells us that developers of AVs should not be aggressive while setting the distance *Threshold of distance for object detection* to a really low value and that AVs need to have some padding while setting safety distance from the OVs in front of it so that even if the ego AV or the OV makes a mistake, the ego AV has some extra safety distance to make up for that mistake.

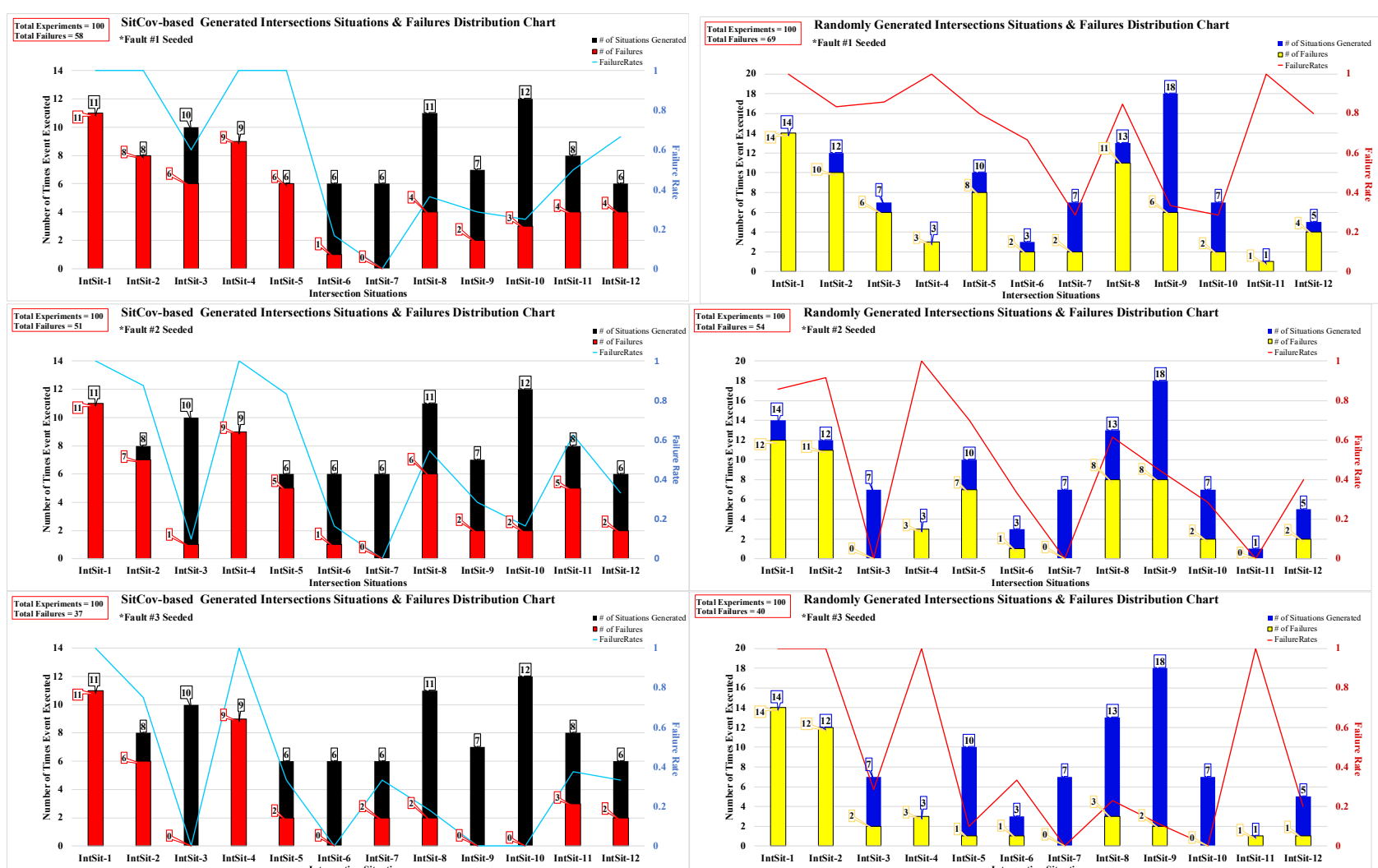

**Fig. 10.** SitCov-based vs Random intersection situation generation & failure distributions, fault-1 seeded.

Table-2 two shows the number of triggered faults from the experiments shown in Fig. 10, by subtracting the total failures of these faults seeded experiments from Fig. 10 by the failures caused when no faults were seeded in experiments shown in Fig. 8, 9.

**Table 2.** Faults Triggered

| Intersection Situation Generation Method | No-Faults Seeded Failures ($f_{NF}$) (100 Runs) | Fault-1 Seeded Failures ($f_{F1}$) (100 Runs) | Fault-2 Seeded Failures ($f_{F2}$) (100 Runs) | Fault-3 Seeded Failures ($f_{F3}$) (100 Runs) | # Times Fault-1 Triggered $= f_{F1} - f_{NF}$ | # Times Fault-2 Triggered $= f_{F2} - f_{NF}$ | # Times Fault-3 Triggered $= f_{F3} - f_{NF}$ |
|---|---|---|---|---|---|---|---|
| Random Generation | 30 | 69 | 54 | 40 | 39 | 24 | 10 |
| Situation Coverage-based Generation | 30 | 58 | 51 | 37 | 28 | 21 | 7 |

Fig. 11 shows the faults triggered by our SitCov AV-testing framework vs random generation from the experiments shown in Fig. 10, and as seen in Fig. 11 the faults being triggered by SitCov-based generation and random generation are somewhat similar in number but the additional benefit that our SitCov-based generation gives us is that the failures rates/fault triggering rates are much more reliable since each situation is thoroughly tested by the SitCov AV-testing framework as compared to random generation where one out of the 12 possible intersection situations (as mentioned in Table-1) could have been tested only one time whereas the other 11 would have been tested 99 times in total, incase of our 100 simulation runs experiments.

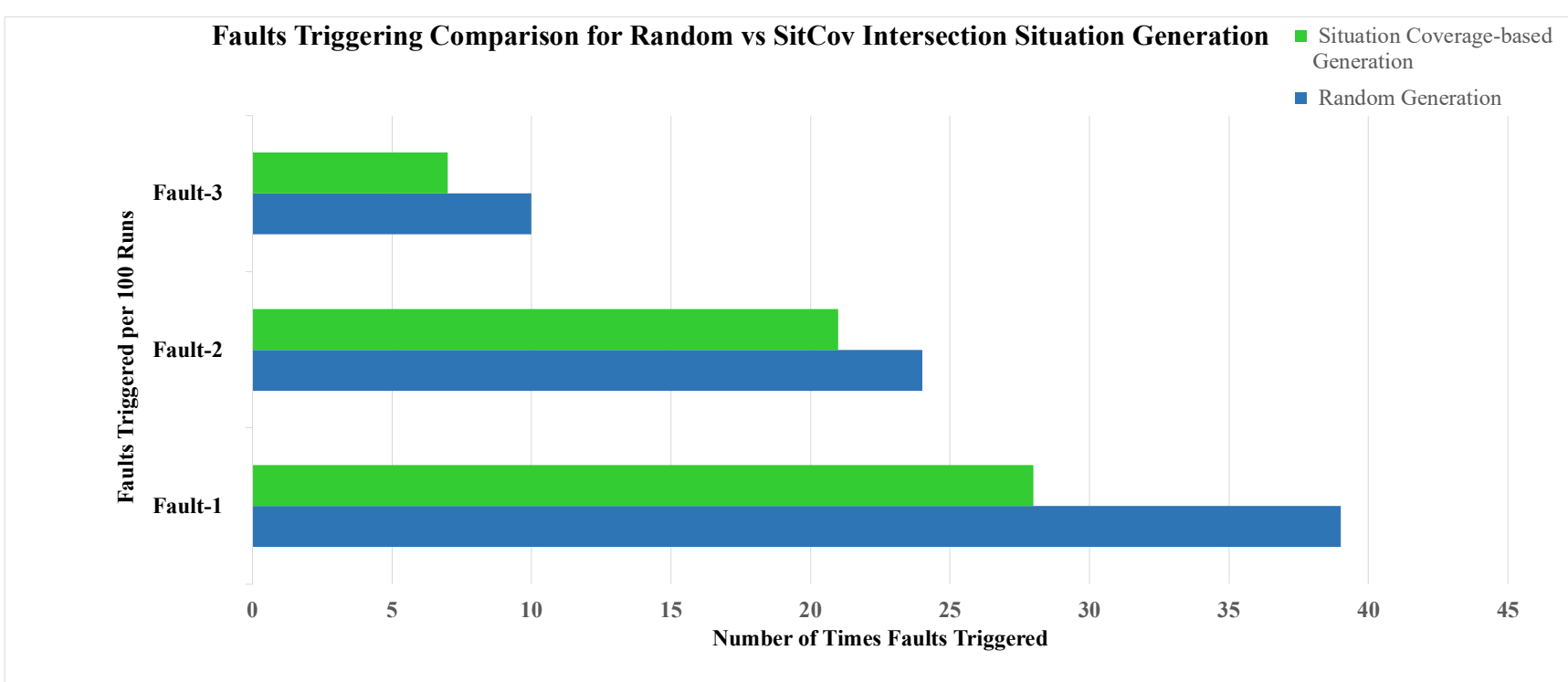

**Fig. 11.** Faults triggering of SitCov-based vs random intersection situation generation

## 4.2 Environmental Conditions Situation Generation Results

We will focus on the environmental conditions situations from the environmental conditions axis (Fig. 2) when comparing our SitCov AV-testing framework with random generation. It is to be noted that actually all of these situation elements from environmental conditions axis and intersection axis are interlinked and correlated as when the discrete situation is generated, it selects some values of all the situation elements from all the axis (environmental conditions axis and intersection axis) from the situation hyperspace, but we are looking at the situation elements separately during our evaluation since analyzing the correlations between the situation elements and their effect on the success/failure of ego AV would be a research project on its own and we definitely recommend it for future work.

**No Faults Seeded Road Friction Experiments.** Fig. 12 shows the results of our SitCov AV-testing frame (in the red and black graph) vs random generation (in the blue and yellow graph) w.r.t road friction, which is a situation element from the environmental conditions axis. Friction values have been divided into 6 bins between 0 and 1 with lowest value being 0.1. The experiments show that random generation as expected has generated highly uneven distribution of friction bins whereas our SitCov AV-testing framework has generated more of an even generation of friction bins. The results from both the graphs in Fig. 12 show us that the failure rate does not get higher as the road friction gets lower, rather the failure rate across all the friction bins is somewhat evenly distributed. This means that this situation element, friction, doesn't affect the success of our ego AV directly that much otherwise it would have had a high failure rate for lower friction values.

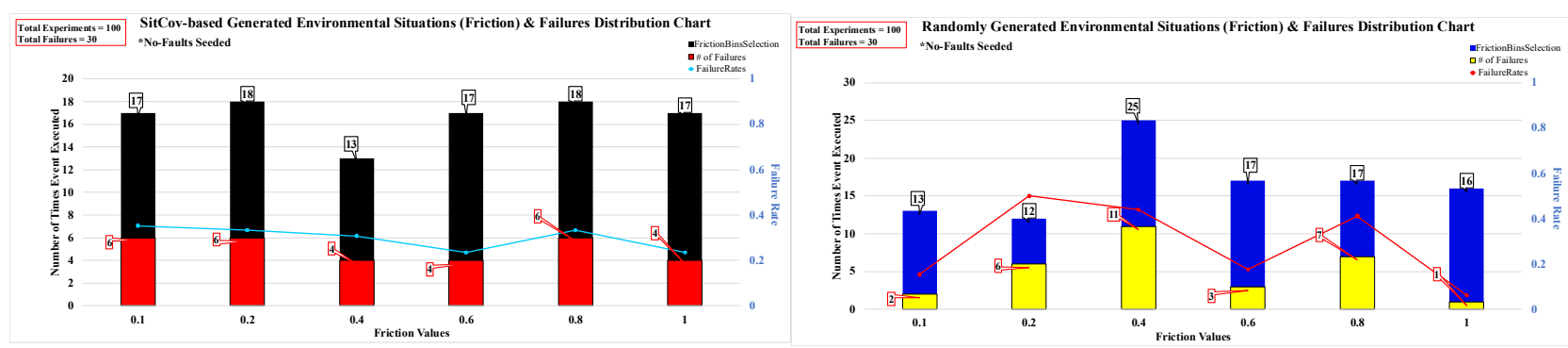

**Fig. 12.** SitCov-based vs Random friction situation generation, no faults seeded.

**Faults Seeded Precipitation Deposits Experiments.** Fig. 12 displays the results of our SitCov AV-testing framework w.r.t the situation element Precipitation Deposits when no seeded faults (top left), fault #1 seeded (top right), fault #2 seeded (bottom left), fault #3 seeded (bottom right).

As seen in Fig. 13, the rise in the failure rate can be seen as the value of precipitation deposits increases in all 4 graphs (no faults seeded and the 3 graphs of the seeded faults). This is due to the fact that our ego AV is relying on the images coming from the dash-cam sensor on the ego AV and uses ML to detect the OV in front of it. The precipitation deposits produce a lot of noise in the incoming images which leads to errors in detecting the OV which causes the ego AV to miss detecting the hazard in front of it, leading to a crash/failure. We can see a big jump in failure rate when fault #1 is seeded, this tells us as we mentioned before that the parameter which fault #1 corresponds to is *Probability of Object Detection Threshold*, and when the image received by the dash-cam on the ego AV is already noisy due to the precipitation deposits, and we introduce the fault #1 which is having a very high detection threshold, our ego AV simply fails to detect the OV in front of it many times and hence has a high failure rate when the precipitation deposits value is high especially when the fault #1 is seeded.

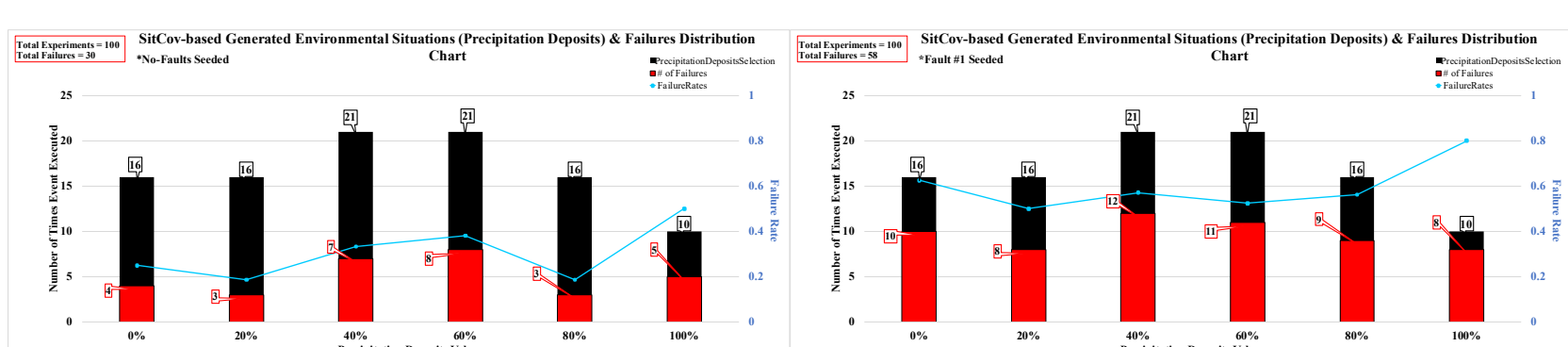

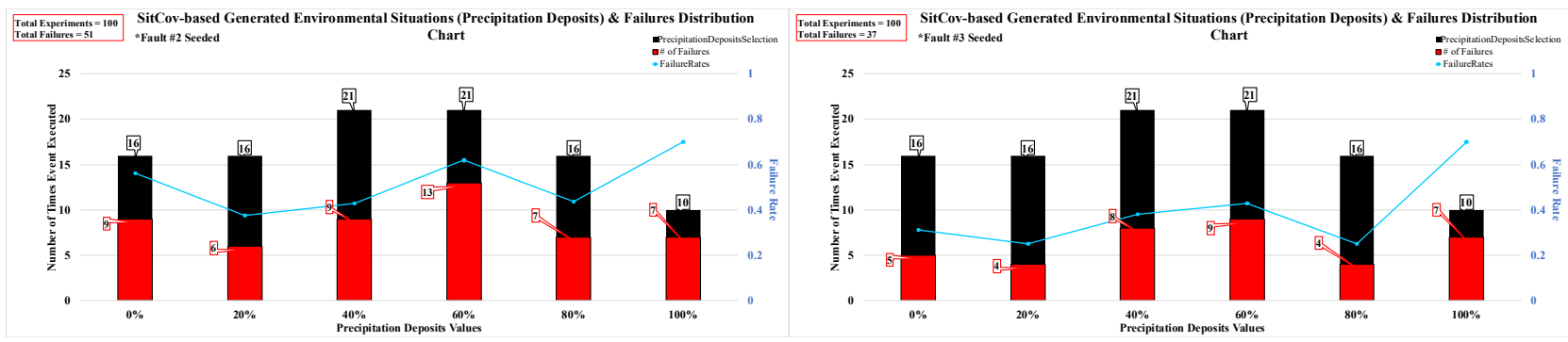

**Fig. 13.** SitCov-based Precipitation Deposits situation generation & failure distributions, no faults seeded.

### 4.3 Summary of Our Results

With respect to RQ1, our experiments suggest that the fault revealing capabilities of our SitCov AV-testing framework vs random generation is more or less the same.

With respect to RQ2, the results produced by the SitCov AV-testing framework are much more convincing than random situation generation, because the results produced by the SitCov AV-testing framework provide additional confidence in the failure rates as the situation coverage-based situation generation of our SitCov AV-testing framework make sure that the situation hyperspace is efficiently covered while also making sure that in case of repetitions of situations, we have an even distribution of repetitions rather than highly varying distributions of situations as seen for random situation generation.

## 5 Conclusion & Future Work

In this paper we have developed a novel situation coverage based (SitCov) AV-testing framework which uses situation hyperspace, a derivative of an ontology that we developed, to automatically generate test suites according to a situation coverage criterion to test AVs for their V&V and safety assurance.

This paper contributes not only theoretically but we have also released our code publicly so researchers can get a head start if they are looking to work in this direction. Our framework has shown that it can find seeded faults and highlight situations with high failure rates and this framework also provides us with confidence in the failure rates of situations it generates due to the even distributions of generating situations, due to the novel mechanism of situation coverage-based situation generation mechanism we have developed for our SitCov AV-testing framework. The developed SitCov AV-testing framework treats AV autonomous driving algorithm as a black-box, hence battling against the AV-IP problem, while testing the AV thoroughly and evenly in all kinds of situations, so that we would have high confidence in the results provided by our SitCov AV-testing framework.

For the future work, we have made our SitCov AV-testing framework really expandable such that many more axis can be added in the situation hyperspace and a lot more dynamic obstacles can be added in the simulation runs instead of just V2V interactions that is currently being done. So, we recommend that the situation hyperspace be expanded more with additional axis, e.g., adding different kinds of roads, adding diverging and merging collisions instead of only cross-collisions with OV, adding pedestrians and cyclists, etc.

We also recommend making use of the failure rate of situations and using that information to skew the generation of next batch of situations in the direction of situation elements that were causing more failure (high failure rate), which can be done using the same mechanism we have developed for doing situation coverage-based situation generation for our SitCov AV-testing framework. Similarly, machine learning can also be added in our SitCov AV-testing framework to generate more interesting situations, i.e., edge-cases, by learning the patterns of combinations of environmental conditions and intersection situations that cause the highest failure rates in the results of our SitCov AV-testing framework test-cases.

*Acknowledgements.* The research presented in this paper has been funded by European Union's EU Framework Programme for Research and Innovation Horizon 2020 under Grant Agreement No. 812.788.